\pdfoutput=1

\documentclass[11pt]{article}

\usepackage[final]{acl}
\usepackage{bm}

\usepackage{times}
\usepackage{latexsym}
\usepackage{float}
\usepackage{hyperref}
\usepackage{url}
\usepackage[T1]{fontenc}
\usepackage{graphics}
\usepackage{graphicx}
\usepackage{caption}
\usepackage{subcaption}
\usepackage{amsmath}

\usepackage[T1]{fontenc}

\usepackage[utf8]{inputenc}

\usepackage{microtype}

\usepackage{inconsolata}

\def\mA{{\bm{A}}}

\def\mM{{\bm{M}}}

\def\mW{{\bm{W}}}
\def\mX{{\bm{X}}}
\def\mY{{\bm{Y}}}

\title{Breaking Symmetry When Training Transformers}

\author{Chunsheng Zuo \\
University of Toronto  \\
\texttt{jason.zuo@mail.utoronto.ca} \\\And
Michael Guerzhoy \\
University of Toronto  \\
\texttt{guerzhoy@cs.toronto.ca} \\}

\begin{document}
\maketitle
\begin{abstract}
The prediction for output token $n+1$ of Transformer architectures without one of the mechanisms of positional encodings and causal attention is invariant to permutations of input tokens $1, 2, .., n-1$. Usually, both mechanisms are employed and the symmetry with respect to the input tokens is broken. Recently, it has been shown that one can train Transformers without positional encodings. This must be enabled by the causal attention mechanism.

In this paper, we elaborate on the argument that the causal connection mechanism must be responsible for the fact that Transformers are able to model input sequences where the order is important. Vertical ``slices" of transformers are all encouraged to represent the same location $k$ in the input sequence. We hypothesize that residual connections contribute to this; we do not find definitive evidence of this.

\end{abstract}

\section{Introduction}


This paper is motivated by recent results~\cite{kazemnejad2023nopelearnspositions, chi2023latent_posinfo_in_nope,haviv2022nope} that indicate that positional encodings are not necessary when training Transformer architectures. We investigate the mechanism through which Transformer architectures are able to obtain position information without positional encoding.

A Transformer architecture without causal attention\footnote{``Causal attention" is the standard term in the literature. ``Causal" to the built-in assumption that ``future" inputs should not affect ``past" inputs.} would be provably equivariant to the permutation of the input tokens~\cite{tsai2019equivariance}, so that the prediction for input token $n+1$ is invariant to permutations of tokens $1, 2, ..., n-1$. Therefore, the causal attention mechanism is required in order for the Transformer to be able to take the order of the input tokens into account.

Our intuition is that residual connections break the symmetry between transformer blocks in different ``vertical slices", so that transformer blocks directly above token number $k$ would tend to contain information related to token number $k$. Our experiments do not provide definitive evidence on whether residual connections help store positional information or merely help with convergence properties.


In our experiments, we use the three-digit addition task. Three-digit addition inherently requires information about the positioning of the input tokens, since, e.g., \texttt{"123+456="} is very different from \texttt{"321+546="}. \cite{lee2023teaching} recently demonstrated a reliable system for training small Transformers from scratch on arithmetic tasks.

Finally, we visualize the correlations between the activations in different layers, which is related to the Transformer's storing positional information.

The rest of the paper is organized as follows. We briefly review attention and causal attention (\ref{back:attention}), residual connections (\ref{back:resid}), and the 3-digit addition task (\ref{back:arithm}. We note that, without a causal attention mechanism, the usual Transformer architecture is equivariant under permutation of the input tokens, and the prediction for token $n+1$ is invariant under permutation of the first $n-1$ tokens (\ref{invariance}). We then empirically investigate Transformer networks trained to perform three-digit addition with some residual connections ablated and report that our Transformers do not converge if enough residual connections are taken out (\ref{ablation}). We investigate the correlation matrices of the activations of our Transformers (\ref{matrices}).

\section{Background}

\subsection{Attention \label{back:attention}}

Mechanisms analogous to modern attention in Transformers have long been used in recurrent neural networks~\cite{bahdanau2014neural}~\cite{schmidhuber1992learning}. An attention mechanism is central to the Transformer architecture~\cite{vaswani2017attention}. 

Given input embeddings $\mX \in \mathbf{R}^{l\times d_{in} }$, a ``non-causal" single-head self-attention mechanism can be formulated as:

\begin{equation} \label{eqn: normal attention weight}
\mA=\frac{(\mX{\mW_Q})({\mX\mW_K})^T}{\sqrt{d_{e}}}
\end{equation}
\begin{equation} \label{eqn: attention out}
\mY=\text{softmax}\left(\mA\right)(\mX\mW_V)
\end{equation}
where $\mA$ is the pre-normalization attention weight matrix, the $\text{softmax}$ applies a row-wise Softmax operation to $\mA$, and $\mY \in \mathbf{R}^{l\times d_{e}}$ is the output of attention. 


The causal attention matrix is as follows.

\begin{equation} \label{eqn: causal attention weight}
\mA_{\text{causal}}=\mA+\mM
\end{equation}
\begin{equation}
\mM_{ij} =
\begin{cases} 
0 & \text{if } j \leq i, \\
-\infty & \text{otherwise}.
\end{cases}
\end{equation}

\begin{figure*}[h]
	\centering
    \includegraphics[width=0.9\textwidth]{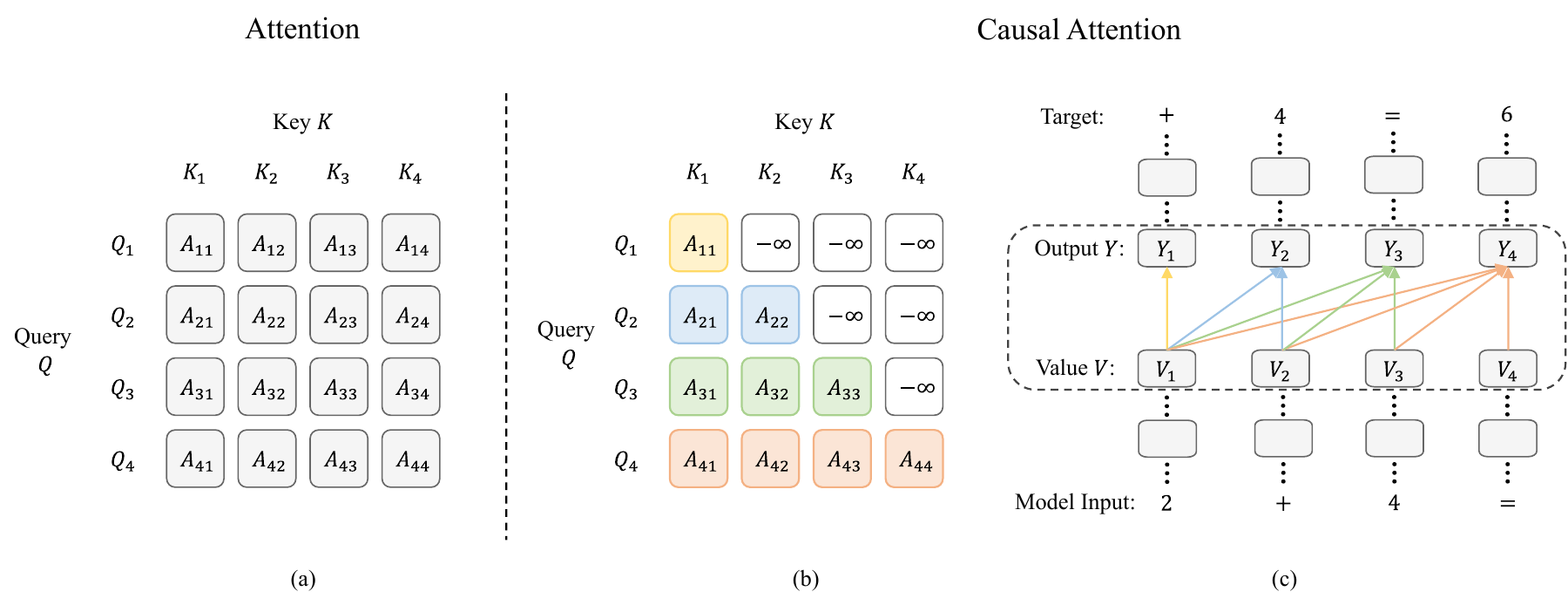}
	\caption{``Non-causal" attention matrix (a), masked attention (b), outputs in an intermediate layer of a transformer computed using masked/causal attention}
	\label{fig:attention}
\end{figure*}

For a block in position $k$, $\mM$ removes the attention weights corresponding to input blocks from the ``future" (i.e., input blocks $k+1, k+2, ..., n$), so that block $k$ is only computed using input blocks $1, 2, ..., k$. Output $Y_k$ is only computed using values $V_1, V_2, ..., V_k$ from the previous layer, and not using $V_{k+1}, ..., V_{n}$, where $n$ is the context window size. See Fig.~\ref{fig:attention}

Note that ``causal attention" is also sometimes used in the context of \textit{generating} output tokens, whereby a new token is generated by only using already-generated tokens. Computationally, this is also accomplished using a masked attention matrix.

\subsection{Residual connections \label{back:resid}}

Residual/skip connections (see, prominently,\cite{he2016deep}, though the idea goes back decades) incorporate the output of layer $L-1$ directly in the output of layer $L$ without intermediate computation. For example, an additive connection might look as follows:
$$O_1 = MLP(Y_1)+\alpha Y_1.$$

Residual/skip connections are thought to address the issue of exploding and vanishing gradients~\cite{he2016deep}. In Transformers, residual connections are thought to be necessary for the Transformer not to degrade very quickly into a rank-1 transformation as the number of layers increases~\cite{dong2021rank_degredation}.

\section{The 3-digit addition task \label{back:arithm}}

In our experiments, we focus on the 3-digit addition task. Essentially, the task involves generating the completion of strings like \texttt{"123+456="}. Following Lee et al., ~\cite{lee2023teaching}, whose code base we also use, we generate the answer in reverse order. The task is selected since the order of the tokens in the task obviously matters a great deal. 

\section{Next-token predictions using ``non-causal attention" are invariant to input permutations \label{invariance}}

\begin{figure}[h]
	\centering

    \centering
    \begin{subfigure}[h]{0.3\textwidth}
        \includegraphics[width=\linewidth]{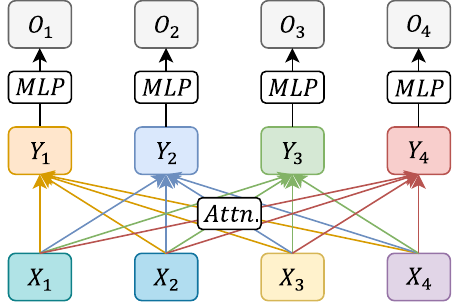}
		\caption{}
		\label{fig:rc_att}
    \end{subfigure}
    \hspace{8mm}
    \begin{subfigure}[h]{0.285\textwidth}
        \includegraphics[width=\linewidth]{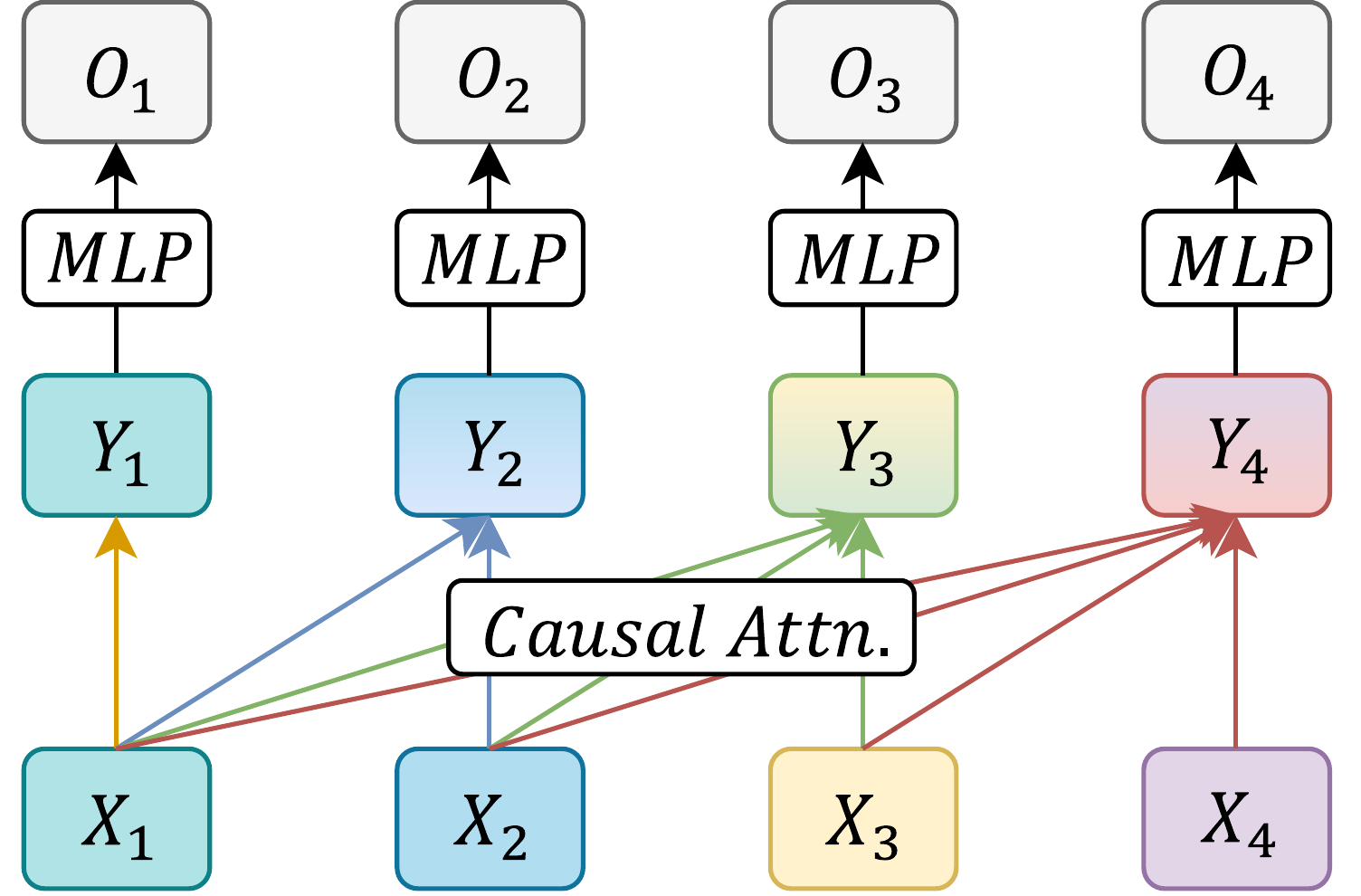}
		\caption{}
		\label{fig:nrc_att}
    \end{subfigure} 
 
	\caption{``Non-causal" attention (a) and causal/masked attention (b)}
	\label{fig:attention_comparison}
 \vspace{-15pt}
\end{figure}

We note that ``non-causal attention" -- attention performed using a non-masked attention matrix -- is inherently invariant to permutations of the input tokens~\cite{tsai2019equivariance}. Consider computing the top-right output in Fig.\ref{fig:rc_att}. Permuting $X_1$ and $X_2$ would simply permute the corresponding attention weights, as well as permute $Y_1$ and $Y_2$, but would not affect the value of $Y_4$. Predictions computed using $Y_4$ (or a block above $Y_4$) would not be affected by the permutation of $X_1$ and $X_2$. More generally, predictions for token $n+1$ would not be affected by permutations of tokens $1, 2, ..., n-1$.

The~\cite{lee2023teaching}, and in most Transformer architectures, the mechanisms that break this symmetry are positional encodings and causal attention. Recent work~\cite{kazemnejad2023nopelearnspositions, chi2023latent_posinfo_in_nope} demonstrates that causal attention is sufficient to break the symmetry.

\section{Some residual connections seem necessary for Transformers to converge \label{ablation}}

In this Section, we report on the empirical observation that, when a sufficient number of residual connections is ablated, the Transformer fails to converge on our task. We speculate that one contributing explanation to that is that Transformers are not able to retain enough information about token positions when too many residual connections are ablated. Some related evidence is in Section~\ref{matrices}.

We train the baseline 6-layer NanoGPT~\footnote{\url{https://github.com/karpathy/nanoGPT}} on the three-digit addition task using learnable absolute positional encoding. We then train it without positional encoding. We then ablate individual residual connections and observe the effect. Our results are summarized in Tables~\ref{t1}~\ref{t2}. We run each configuration 5 times. We obtain nearly-perfect performance both with and without positional encodings (``NoPE"). Convergence something suffers when 2 residual connections are removed, although the model sometimes converges. We are not able to get the model to converge after ablating three consecutive residual connections.

Note that each layer actually has two residual connections: input to pre-MLP and pre-MLP to output. When we ablate from layer L, we ablate both connections.

Although positive convergence results prove that the model can converge, negative results might simply indicate that we have not found the right hyperparameters or have not trained for long enough. However, we obtain strong evidence that, at least as far as convergence is concerned, removing enough residual connections hurts performance.

\begin{table*}[h]
\caption{Three-digit addition performance (in \%) performance after removing residual connection (RC) from 0 or 1 layers}
\vspace{3pt}
\renewcommand{\arraystretch}{1.2} 
\label{t1}
\centering
\begin{tabular}{c|ccccccc}

Layers without RC & \{\}   & \{1\} & \{2\} & \{3\} & \{4\} & \{5\} & \{6\} \\
\hline
Original (avg.) & 100.00 & 99.97 & 99.84 & 99.51 & 99.75 & 99.86 & 99.90 \\
NoPE (avg.) & 99.59 & 96.96 & 95.46 & 89.83 & 69.13 & 95.99 & 99.48 \\
\end{tabular}
\end{table*}

\begin{table*}[h]
\caption{Three-digit addition performance (in \%) performance after removing residual connection (RC) from 2 or 3 layers}
\vspace{3pt}
\renewcommand{\arraystretch}{1.2} 
\label{t2}
\centering
\begin{tabular}{c|ccccccccc}

Layers without RC & \{1,2\} & \{2,3\} & \{3,4\} & \{4,5\} & \{5,6\} & \{1,2,3\} & \{2,3,4\} & \{3,4,5\} & \{4,5,6\}\\
\hline
Original (min) & 98.53 & 0.01 & 0.01 & 0.01 & 0.01 & 0.01 & 0.01 & 0.01 & 0.01 \\
NoPE (min) & 10.36 & 0.01 & 0.01 & 0.01 & 0.01 & 0.02 & 0.01 & 0.01 & 0.01 \\
Original (max) & 99.74 & 90.75 & 2.52 & 99.98 & 99.62 & 0.82 & 0.02 & 0.03 & 0.02 \\
NoPE (max) & 80.12 & 0.15 & 0.07 & 0.07 & 0.69 & 0.13 & 0.09 & 0.03 & 0.04 \\
\end{tabular}
\end{table*}

\begin{figure*}[ht]
	\centering
    \includegraphics[width=0.99\textwidth]{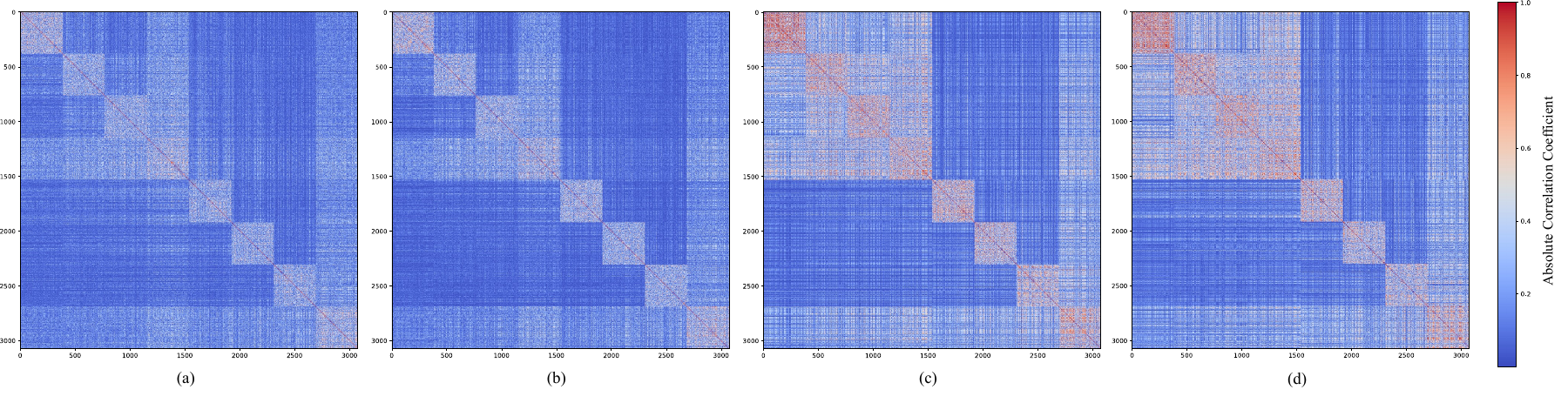}
	\caption{Absolute value of the correlation matrices for output embeddings from layer 1 of NoPE models with residual connections removed at blocks \{\} (a) \{0\} (b) \{0,1\} (c), and \{0,1\} with a different random initialization (d). Typical results. Note the fact that there are more off-diagonal and off-block-diagonal large values without residual connections. More results in Figs.~\ref{fig:matrices1}~\ref{fig:matrices2}.
 }
	\label{fig:matrices0}
\end{figure*}

\begin{figure*}[ht]
	\centering
    \includegraphics[width=0.99\textwidth]{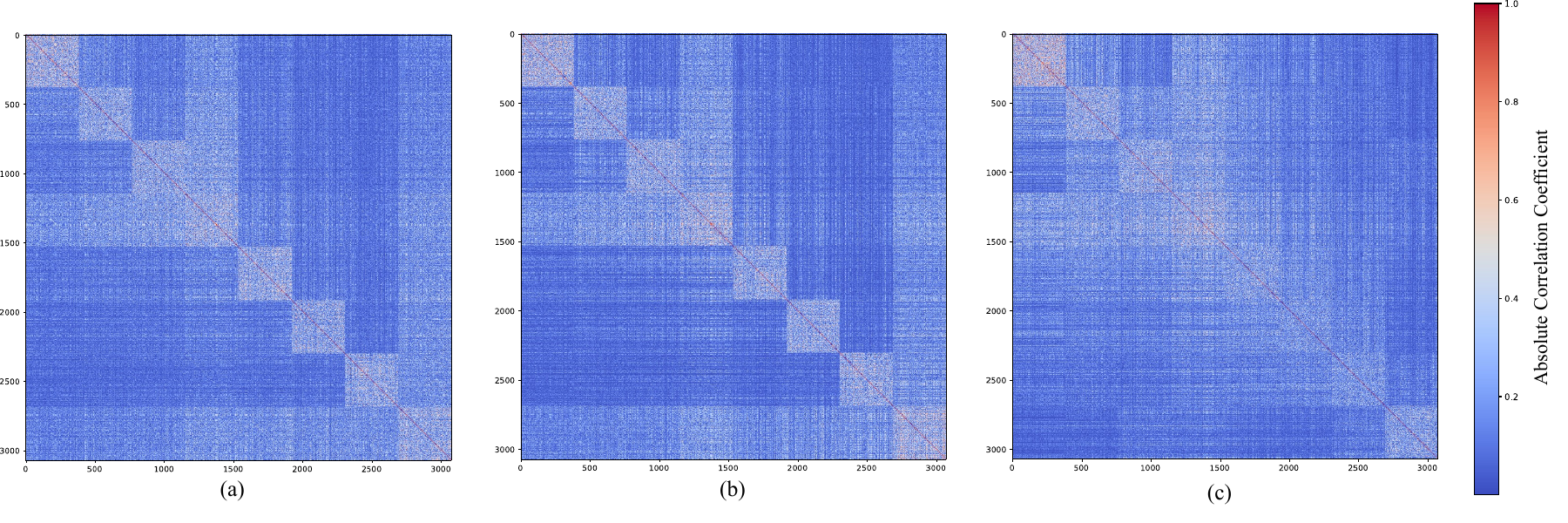}
		\caption{Absolute value of the correlation matrices for output embeddings from layer 1 (a), 3 (b), and 6 (c) of NoPE models with no residual connections removed. 
 }
	\label{fig:matrices1}
\end{figure*}

\begin{figure*}[ht]
	\centering
    \includegraphics[width=0.99\textwidth]{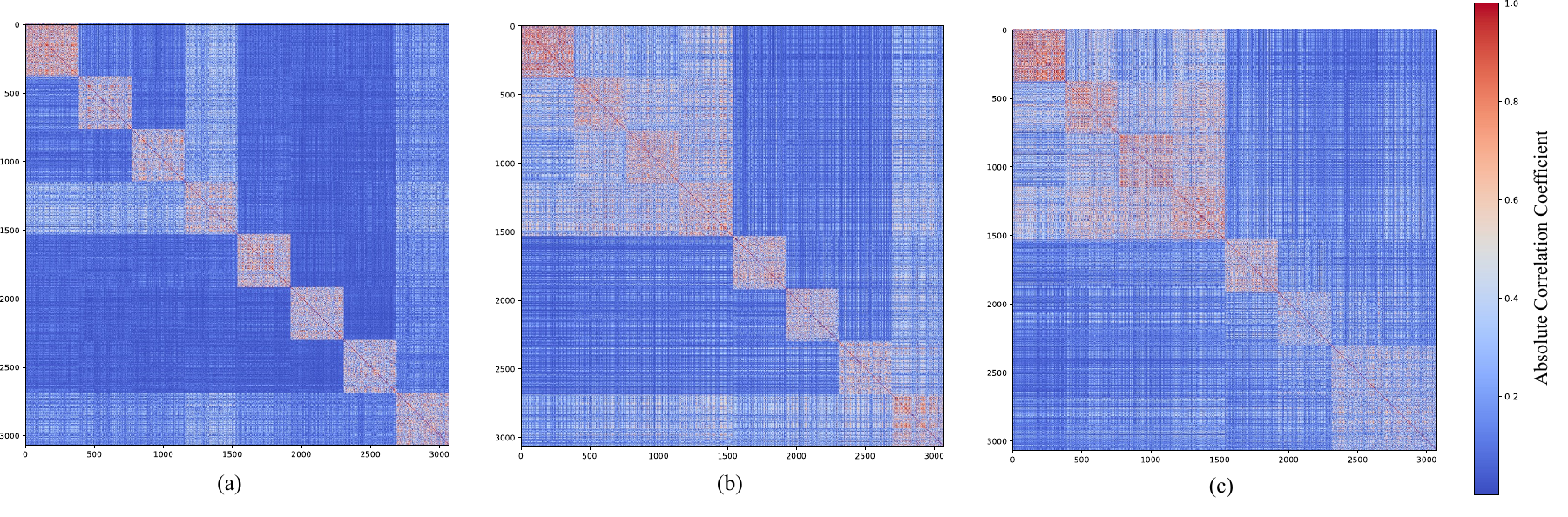}
	\caption{Absolute value of the correlation matrices for output embeddings from layer 1 (a), 3 (b), and 6 (c) of NoPE models with residual connections removed at layer {0,1}. 
 }
	\label{fig:matrices2}
\end{figure*}

\section{Correlations between activations  \label{matrices}}
Transformers are known to keep information about token $k$ in the $k$-th column of the transformer block. For example, probing of language models~\cite{hewitt-liang-2019-designing} relies on this fact. 

As shown in Fig.~\ref{fig:matrices0}, we demonstrate a visualization of the absolute value of the Pearson correlations between all the activations in a layer of our Transformer trained on the three-digit addition task.

We flatten the activations of the Transformer into a 1-D vector by rasterizing all the activations in row-major order. Activations from the same attention block in the same layer are rasterized to nearby coordinates.

The ``blocky" structure indicates that, within each block, activations can get ``permuted" to some extent layer-to-layer. Activations that belong to the same block in the same layer are likely correlated. If the transformer ``permutes" the location where information about token $k$ is stored between layers $l_1$ and $l_2$, we'd expect to see an off-diagonal block with high correlations, which we sometimes observe.

The observations that there are more pronounced ``off-diagonal" blocks when there are fewer residual connections indicate that residual connections play a role in keeping information from token $k$ in the k-th vertical slice of the transformer.


\section{Conclusions}
In a no-positional-encodings setting when training Transformers, causal attention is necessary. Residual connections play a role in improving convergence. Although there is a theoretical reason to believe they would help with preserving positional information, we do not have definitive evidence of that. In future experiments, we will attempt to investigate ablating the possible role of the residual connections in preserving position information while keeping their role in improving convergence properties.

\bibliography{custom}


\end{document}